\documentclass[10pt,twocolumn,letterpaper]{article}

\usepackage{iccv}
\usepackage{times}
\usepackage{epsfig}
\usepackage{graphicx}
\usepackage{amsmath}
\usepackage{amssymb}

\usepackage{booktabs}
\usepackage{adjustbox}
\usepackage{multirow}
\usepackage{float}
\usepackage{subcaption}
\usepackage[table,xcdraw]{xcolor}
\usepackage[breaklinks=true,bookmarks=false]{hyperref}

\iccvfinalcopy % *** Uncomment this line for the final submission

\def\iccvPaperID{6574} % *** Enter the ICCV Paper ID here

\begin{document}

\title{
% RotaTR: Save the Transformer for Dense and Rotated Object Detection.
RotaTR: Detection Transformer for Dense and Rotated Object
}

\author{
Yuke Zhu, Yumeng Ruan, Lei Yang, Sheng Guo\\
{Mybank Technology\hspace{0.5cm}}\\
{\texttt{\small{\{felix.yk, guosheng.guosheng\}@mybank.cn}}}
}

\maketitle

\newcommand{\FFN}[0]{\text{FFN}}

%%%%%%%%% ABSTRACT
\begin{abstract}
   Detecting the objects in dense and rotated scenes is a challenging task. Recent works on this topic are mostly based on Faster RCNN or Retinanet. As they are highly dependent on the pre-set dense anchors and the NMS operation, the approach is indirect and suboptimal.
   The end-to-end DETR-based detectors have achieved great success in horizontal object detection and many other areas like segmentation, tracking, action recognition and etc. 
   However, the DETR-based detectors perform poorly on dense rotated target tasks and perform worse than most modern CNN-based detectors. In this paper, we find the most significant reason for the poor performance is that the original attention can not accurately focus on the oriented targets.  Accordingly, we propose Rotated object detection TRansformer (RotaTR) as an extension of DETR to oriented detection. 
   Specifically, we design Rotation Sensitive deformable (RSDeform) attention to enhance the DETR's ability to detect oriented targets.  It is used to build the feature alignment module and rotation-sensitive decoder for our model. We test RotaTR on four challenging-oriented benchmarks. It shows a great advantage in detecting dense and oriented objects compared to the original DETR. It also achieves competitive results when compared to the state-of-the-art. The code will be released.
   
\end{abstract}

%%%%%%%%% BODY TEXT

\begin{figure*}[!ht] 
\centering 
\includegraphics[width=\textwidth]{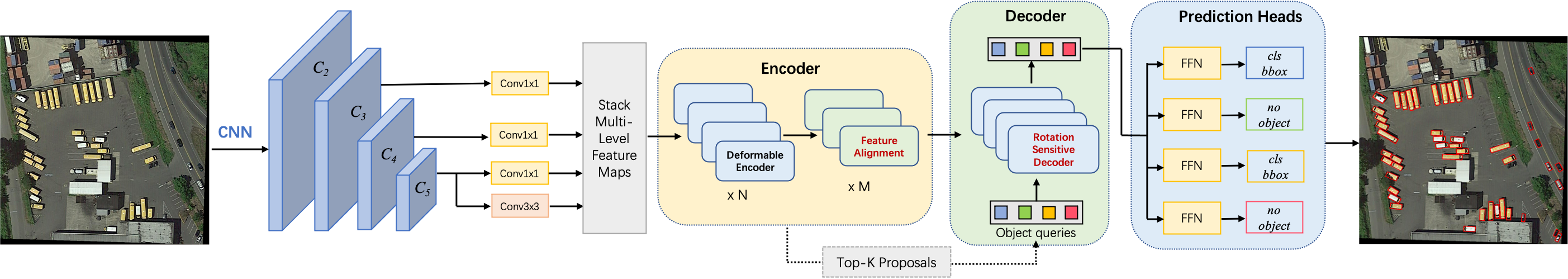} 
\captionsetup{font={small}}
\caption{A schematic overview of RotaTR. Note that the RotaTR can be used in the two-stage form and one-stage form. When in two-stage form, the encoder will generate top-k proposals as the initialization for the decoder query. }
\label{fig:structure}
\vspace{-10pt}
\end{figure*}

\section{Introduction}
\label{sec:intro}
% intro of rdet and its applications
% Most, faster rcnn, retinanet, relies on pre-set anchors, NMS post proc tedious. drawback, horizontal trend, O2DETR,
% our work
%contribution
Oriented object detection is a more general and challenging task than conventional horizontal detection. It has a wide range of applications in scene text detection\cite{karatzas2015icdar}\cite{yao2012msra_tda}, aerial object detection\cite{xia2018dota}\cite{liu2017hrsc}, faces and retail scenes\cite{shi2018face}\cite{chen2020piou}. 
% Compared with the conventional horizontal detection,it is a more challenging task as it has to precisely locate the arbitrarily oriented target using a rotated bounding box. 
% Although oriented detection has been studied for years,

Nowadays, most state-of-the-art methods\cite{R3Det}\cite{S2ANet}\cite{ding2019roitrans}\cite{orcnn} on oriented detection are based on Faster RCNN\cite{fasterrcnn} or Retinanet\cite{2017Focal}.
% Most work on oriented object detection are based on the Faster RCNN\cite{fasterrcnn} or Retinanet\cite{2017Focal}. For example, modern one stage oriented object detectors R3Det\cite{R3Det}, S2ANet\cite{S2ANet} introduce the feature refinement for Retinanet. Two stage detectors RoI-Transformer\cite{ding2019roitrans}, Oriented RCNN\cite{orcnn} improve the oriented region proposal for Faster RCNN. 
They highly depend on pre-set dense anchors for box regression and non-maximum suppression (NMS) for duplicate removal. These two hand-crafted modules lead to a tedious and redundant detection process, which is an indirect and sub-optimal solution to the oriented detection problem.
In contrast, the idea of end-to-end object detection has been favored by many researchers in conventional detection. A lot of methods have been proposed to design an end-to-end detector from the perspective of anchor free\cite{Huang2015DenseBoxUL}\cite{Redmon2016YouOL}\cite{Law2019CornerNetDO}\cite{Tian2019FCOSFC}, NMS free\cite{hu2018relationnet}\cite{duan2019centernet}\cite{onenet}\cite{wang2021end}\cite{sparsercnn} or the recent transformer\cite{detr}\cite{anchordetr}\cite{dabdetr} structure.
% Nowadays, in horizontal detection, many researchers are devoted to develop the end-to-end detection pipeline, in which all modules are differentiable so that the detector can be optimized as a whole. 
% The research direction can be summarized into two groups: designing anchor-free detectors\cite{duan2019centernet} and nms-free detectors\cite{duan2019centernet}.
However, in oriented detection, there are few works addressing this issue. 

Recently,  DETR\cite{detr} and its followers\cite{deformdetr}\cite{anchordetr}\cite{dabdetr} bring new insight into the detection paradigm. 
The DETR regards object detection as a sequence-to-sequence problem. It uses a set of object queries to probe into the image context and directly generates a set of predictions without the need for NMS. 
The new paradigm of DETR is general and has been successfully adapted to many other areas like instance segmentation\cite{Zheng2021RethinkingSS}\cite{Cheng2021PerPixelCI}\cite{Cheng2022MaskedattentionMT}, action recognition\cite{Xu2021LongST}\cite{liu2022end}\cite{Shi2022ReActTA}, multi-object tracking\cite{Wang2021EndtoEndVI}\cite{Hwang2021VideoIS}\cite{Wu2021SeqFormerAF} etc.
So, it is a natural thought that DETR-like structures can also be easily adapted to oriented object detection. 
Based on the idea, we design a simple oriented detector Deformable-DETR-O based on Deformable DETR\cite{deformdetr} (detailed in Sec-\ref{sec:adapt}) and test it on the common oriented benchmark.
% and compare it with other CNN-based methods in the dense and oriented scenery. 
% The adaptation details of Deformable DETR-O is introduced in Sec-\ref{sec:adapt}. 
However, the results are frustrating. The adapted detector performs even inferior to Retinanet-O, which usually serves as the baseline method for most oriented detection methods (Shown in Table-\ref{tab:dota_comp}). The poor performance is especially significant for densely located targets. The phenomenon can also be supported by some observations from DAB-DETR\cite{dabdetr}, which reports a lot of failure cases in dense scenes.
The recently published O2DETR\cite{o2detr} is the first attempt to design DETR-like oriented detector. It also reports low performance on such occasions. 
So far, no other work has attempted to explain the poor performance of DETR-like structures for oriented and densely located targets.
To this end, we take a closer look at the cross-attention module in the transformer decoder. Specifically, the sampled locations of the deformable cross attention are visualized. The result shows that the sampled locations of the deformable attention (Shown in Figure-\ref{fig:pts_plot}(d) can hardly be aligned with the target geometry. 
It suggests that it is difficult for deformable attention to dynamically attend to arbitrarily oriented targets.

In this paper, we propose Rotated object detection TRansformer (RotaTR) as a solution to extend DETR to oriented object detection. 
Motivated by the above observation, we design Rotation Sensitive Deformable (RSDeform) attention to enhance the deformable attention with the ability to learn orientations. In RSDeform, the orientation, as well as the bounding box's geometry, are explicitly exploited to guide the attention sampling. 
The RSDeform forms the basic building block for the decoder. Besides, we also use it to build a feature alignment module to solve the feature misalignment problem caused by arbitrary orientation. 
The RSDeform introduces no extra parameters and computation for oriented sampling can be neglected. 
% With the help of RSDeform, the RotaTR greatly complements the shortcoming of DETR in detecting dense and oriented objects. 
Besides, we also follow the DAB-DETR\cite{dabdetr} to introduce the concept of dynamic anchors to simplify the query design.
We also notice that the regression of the rotated bounding box may encounter the problem of ambiguity and in-continuity. The regression issue has been discussed by a large number of researchers\cite{Yang2019SCRDetTM}\cite{Qian2022RSDetPM}. Inspired by previous work, we design a simple point set loss to calculate the distance between the predicted and target point set. The point set loss is simple and intuitive and proves effective in our experiments.

In summary, the contributions of this work include: 
(1) We propose RotaTR as an extension of DETR to oriented detection. It greatly strengthens the DETR's ability in capturing arbitrary orientations. 
(2) We design the RSDeformable attention to facilitate orientated attention. It is an individual module and can be applied to other structures.
(3) We propose point set loss as a simple oriented regression loss. It proves effective in both RotaTR and other oriented-detectors.

\section{Related Work}
\label{sec:related}
\subsection{End-to-End Object Detection}
End-to-end object detection refers to the detection system that is free of any non-differentiable modules. 
Early methods towards end-to-end detection are based on CNN detectors. Most of them aim to eliminate the anchors\cite{Tian2019FCOSFC}\cite{Redmon2016YouOL}\cite{pan2020dynamic} or nms\cite{hu2018relationnet}\cite{duan2019centernet} operation.
% To our knowledge, \cite{stewart2016end} is the first attempt towards designing an end-to-end deep learning-based detector. 
% They use encoder-decoder structure to build detection as a sequence-to-sequence problem and adopt bipartite matching to assign the target labels. This work has now proven to be one of the most effective way towards end-to-end detection. Unfortunately, as the concurrent Faster RCNN\cite{fasterrcnn} has overwhelmed the entire detection area, this work has almost been emerged.
% Latter work towards end-to-end detector are mostly build upon the Faster RCNN style. For these detectors, eliminating the hand-crafted Non-Maximum Suppression(NMS) is becoming the primary goal.
% Relation network\cite{hu2018relationnet} proposes a duplicate removal module that learns to remove redundant detections. CenterNet\cite{duan2019centernet} proposes to directly detect the object center keypoints.

Recently, DETR\cite{detr} proposes a transformer-based end-to-end detector.
Since its publication, it has attracted great attention and a lot of successors\cite{deformdetr}\cite{anchordetr}\cite{conditionaldetr}\cite{dabdetr} have been proposed. Later, some CNN-based methods\cite{onenet}\cite{wang2021end}\cite{sparsercnn} also followed a similar idea of DETR to design the NMS-Free detector.

% OneNet\cite{onenet} introduced classification cost to one-stage detector in order to produce one-to-one prediction without the NMS procedure. SparseRCNN\cite{sparsercnn} designed a fixed sparse set of learned object proposals to replace the Region Proposal Network(RPN)\cite{fasterrcnn}. 

% As the multi-head self-attention architecture of transformers becomes more and more popular in computer vision, DETR\cite{detr} and its several extended versions are proposed. DETR directly predicts the object boxes by combining a CNN backbone with a transformer encoder-decoder architecture. It utilizes  bipartite matching to make predictions uniquely assigned with ground truth during training. However, DETR suffers from slow convergence and limited feature spatial resolution. Deformable-DETR\cite{zhu2020deformable} restricts each object query to a small set of key sampling points around a reference, which can definitely accelerate convergence and achieve better performance. Conditional-DETR\cite{meng2021conditional} presents a conditional cross-attention mechanism, which can localize the distinct regions for object classification and box regression through the conditional spatial query. Anchor-DETR\cite{wang2022anchor} proposes to design multiple pattern object queries based on anchor points and design an attention variant, which can reduce the memory cost. DAB-DETR\cite{liu2022dab} uses anchor size to modulate the positional cross-attention map in Transformer decoders and perform dynamic anchor update layer by layer.

In contrast, for oriented object detection, the efforts toward end-to-end detection are nearly absent. The recent work O2DETR\cite{o2detr} first attempts to introduce the DETR into oriented detection. However, the poor performance of DETR in dense scenes is not solved. The final performance lags far behind most of the SOTA CNN-based methods. 

% The idea of abandoning NMS and embracing end-to-end system in horizontal object detection has been widely accepted. But in more general arbitrary pose object detection, end-to-end work is almost absent. Oriented RepPoints\cite{reppoints} proposes an effective adaptive points learning approach to aerial object detection by taking advantage of the adaptive points representation. O2DETR\cite{o2detr} is the first attempt to apply transformer to the oriented object detection task, which utilizes separable depthwise convolutions to replace attention mechanism. 

\subsection{Feature Misalignment in Oriented Detection}
In object detection, feature alignment refers to the spatial alignment between image features and anchor boxes.
For example, Mask RCNN\cite{He2017MaskR} proposes RoIAlign, which uses bi-linear interpolation to replace max-pooling to avoid the quantization error in RoIPooling\cite{fasterrcnn}, thus leading to better alignment between extracted features and anchors. Following work like Guided Anchoring\cite{GuidedAnchoring}, AlignDet\cite{AlignDet} also seek to better align features with anchors for accurate feature extraction.
% Feature alignment, which usually refers to the alignment between convolution features and anchor boxes, is such a crucial thing in arbitrary pose object detection that lots of work put it in a core position. 

% Early work on horizontal object detection has begun to try to do feature alignment. In two-stage detectors, an RoIPooling operator is introduced to  extract fixed-length features of sub-regions which should be focused. On the basis of RoIPooling, RoIAlign uses bi-linear interpolation to replace max-pooling to avoid the the quantization error and Deformable RoIPooling proposes an adaptive offset to each sub-region, which brings better performance. Besides, Guided Anchoring\cite{GuidedAnchoring} proposes a new anchors generation method which predicts sparse and arbitrary shaped anchors to make them better matched with feature map. AlignDet\cite{AlignDet} devises RoIConv, which aligns features and corresponding anchor boxes in a principled way. These 
% works has achieved certain results in normal horizontal object detection scenarios. 

% However, different from the horizontal object detection task, oriented object detection relies on oriented bounding boxes (OBB) to capture the arbitrarily oriented objects. Consequently,
For oriented detection, the introduction of angle dimension makes feature alignment more important.
% misalignment between objects and anchors becomes more significant and complicated. 
% In order to reduce the angle mismatch and improve the performance, some special designed architectures are proposed. For example, 
For better feature alignment, R3Det\cite{R3Det} adds Feature Refinement Module to reconstruct the feature map. S2ANet\cite{S2ANet} designs Feature Alignment Module and Oriented Detection Module to generate high-quality anchors and aligned features. Oriented RCNN\cite{orcnn} uses Rotated RoIAlign to extract rotation-invariant features from oriented proposals generated from oriented RPN. 

% Inspired by so much great work mentioned above, our work will also consider feature alignment as an important part.
In this work, we also find that the core reason for the poor performance of DETR in oriented detection is the lack of alignment ability. The proposed RotaTR uses specially designed RSDeform attention to solve this problem.

\section{Method}
\label{sec:method}
\subsection{Adapting DETR for Rotated Object}
\label{sec:adapt}
% angular regression
% adapting to deformable detr
% optimization, poly iou loss + L1 loss
The first step to adapt DETR to rotated detection is to add an extra channel for the angle regression. 
Since two quadrants are enough to represent any rotated box, we restrict the angle to a certain range, e.g. $[-\pi/2, \pi/2]$, to remove its periodicity. Given a decoder embedding $x$,  the predicted rotated box is computed as:
\begin{align}
    (x, y, w, h, \hat{\theta}) &= \sigma(\FFN(x))
\end{align}
where $\hat{\theta}$ is the intermediate output for the angle without unit. $\sigma$ is the sigmoid operation. The final angle is got by $\theta = \hat{\theta}A - A/2$ where $A$ is the pre-defined angle range. 

As the original DETR converges slowly and performs poorly compared with its successors, we choose the Deformable-DETR as our baseline method. Similar to Deformable-DETR, the regression target is defined as the increment with respect to the previous prediction $\mathbf{p} = (p_x, p_y, p_w, p_h, p_\theta)$. Given the regression $(\delta \hat{x}, \delta \hat{y}, \hat{w}, \hat{h}, \hat{\theta}) = \FFN(x)$,  the final predicted bounding box $(x,y,w,h,\theta)$ is computed as
\begin{equation}
\begin{aligned}
    (x, y) &= \left(\sigma(\delta \hat{x} + \sigma^{-1}(p_x)), \sigma(\delta \hat{y} + \sigma^{-1}(p_y))\right) \\
    (w, h) &= \left(\sigma(\delta \hat{w} + \sigma^{-1}(p_w)), \sigma(\delta \hat{h} + \sigma^{-1}(p_h))\right) \\
    \theta &= A\sigma(\delta \hat{\theta} + \sigma^{-1}(p_\theta)) - A/2
\end{aligned}
\vspace{-6pt}
\end{equation}
where $\sigma^{-1}$ is the inverse sigmoid function. 

The optimization process is kept identical to the Deformable DETR, which uses the combination of Focal loss, L1 loss, and IoU loss as the supervision and utilizes the Hungarian matching algorithm to assign labels. The difference is that we use the Rotated-IoU loss to substitute the horizontal IoU loss. Besides, the L1 loss is applied to 5-D rotated boxes. 

\subsection{Overall Architecture}
% based on dab detr with deformable attention
% dynamic anchor for rotated box
% angular encoding in self attention
% rotation sensitive cross attention with dynamic anchor modulation
% Following DETR, our model is an end-to-end detector which includes a backbone, Transformer encoders and decoders, and prediction heads for boxes and labels. 
\textbf{Structure Overview.} The structure of RotaTR is shown in Figure-\ref{fig:structure}. Following recent work on improving DETR, we design RotaTR based on two main ideas. Specifically,  we use dynamic anchor\cite{dabdetr} to simplify the query design and multi-scale deformable attention\cite{deformdetr} for faster convergence.
% we take the idea of dynamic anchor from DAB-DETR and multi-scale deformable attention from Deformable DETR to design the dab-deformable DETR as our model's infrastructure. 

% The inference pipeline is as follows. Given an image, we use the CNN backbone to extract multi level features. Then, we flatten and stack the features and feed them to the transformer encoder to generate refined features. 
% Finally, in the decoding process, a set of object queries will be passed to the decoder. Each of them is responsible for an instance detection.

% The encoder is composed by several regular encoder layers and 
The encoder shares a similar structure with Deformable-DETR. The only difference is that we replace the last $M$ layers with feature alignment layers. We will introduce it in Sec-\ref{sec:feature_align}.
The decoder is a stack of rotation-sensitive decoder layers. Each layer is composed of a regular multi-head self-attention, a rotation-sensitive deformable (RSDeform) cross-attention, and a feed-forward layer.

One point worth noting is that RotaTR also has a two-stage form like Deformable-DETR. In the two-stage form, the decoding queries are initialized by the top-k selected proposals generated by the encoder output. We refer the readers to \cite{deformdetr} for details.

The key component of both the alignment layer and RSDeform Attention layer is the RSDeform attention, which learns to explicitly modulate the sampling locations based on the given reference anchors and aggregate the sampled features. It is detailed in Sec-\ref{sec:rsdeform}.

\textbf{Query Design.} We follow the idea of DAB-DETR\cite{dabdetr} to design the dynamic-oriented anchor and use it to form our query. Given a dynamic anchor box $A_q = (x_q, y_q, w_q, h_q, \theta_q)$, its positional query $p_q$ is generated by:
\begin{equation}\label{eq:pos_q}
    p_q = \text{MLP}(\text{PE}(A_q))
    \vspace{-5pt}
\end{equation}
where \text{PE} means the operation to generate sinusoidal embeddings from float numbers. It encodes both the coordinates of anchor boxes and the angular information: 
\begin{equation}
\begin{aligned}
    \label{eq:pos_enc_q}
    \text{PE}(A_q) = \text{CAT}(\text{Pe}(x_q), \text{Pe}(y_q)&,\text{Pe}(w_q), \text{Pe}(h_q), \\ 
    &\sin(\theta), \cos(\theta))
\end{aligned}
\vspace{-5pt}
\end{equation}
where \text{CAT} means the tensor concatenation operation and \text{Pe} is the sinusoidal generation function identical to the that in DETR\cite{detr}. 

In self-attention, the query, key, and value embeddings are defined by:
\begin{equation}
\vspace{-5pt}
    \begin{aligned}
    \label{eq:self}
    \text{Query, Key} &:= c_q + p_q\\
    \text{Value}&:= c_q
    \end{aligned}
\end{equation}
where $c_q$ represents the content embedding (decoder embedding).
In cross attention, it is similar to that in Deformable DETR except that we change the reference point to a dynamic anchor $A_q$ and change deformable attention to RSDeform attention.

\subsection{Rotation Sensitive Deformable Attention}
\label{sec:rsdeform}
For standard multi-scale deformable attention, it calculates the sampled locations for each reference anchor and then samples over the input multi-level features and sums the sampled features using learned attention weight. Given with a single-level feature map $x \in \mathbb{R}^{C \times H \times W}$, query embedding $z_q$, and its reference anchor $A_q = (\mathbf{p_q}, w_q, h_q, \theta_q)$ with $\mathbf{p_q}=(x_q, y_q)$, the deformable attention in a single head is calculated by
\begin{equation}
    \text{Deform}(z_q, A_q, x) = \sum_{k=1}^{K}{W_{qk} W^{'} x(\mathbf{p_q} + \Delta \mathbf{p_{qk}})}
\end{equation}
where $k$ indexes the sampled keys and $K$ is the total number of keys. $W^{'}$ is the projection weight for input image features. $W_q \in \mathbb{R}^{1 \times K}$ is the attention weight calculated by $W_q = \text{MLP}(z_q)$, and $W_{qk}$ is its $k$-th component. $\Delta \mathbf{p_{q}}$ is the sampling offsets obtained by $\Delta \mathbf{p_{q}} = \text{MLP}(z_q)$. 

% Compared with the standard deformable attention, the RSDeform attention explicitly models the influence of rotation and anchor geometry on the sampling locations. 
RSDeform attention is designed on the basis of two considerations. 
Firstly, the learned offsets $\Delta \mathbf{p_q}$ in the original deformable attention are forced to implicitly adapt to the angular variation. 
% While we believe that the direct modeling can help ease the burden of learning and get better performance. 
While experimentally, we find the adaptability of offsets to orientation is weak. 
Thus, for each reference anchor $A_q$, 
the modulated offset $\mathbf{\tilde{p}_q}$ is explicitly calculated by 
\begin{equation}
    \Delta \mathbf{\tilde{p}_q} = \Delta\mathbf{p_q} \cdot R^T(\theta)
\end{equation}
where $R(\theta) = (\cos\theta, -\sin\theta; \sin\theta, \cos\theta)^T$ is the rotation matrix. Secondly, the learned offsets $\Delta \mathbf{p_q}$ in the original deformable attention are not imposed by any restriction. The observed sampling locations are usually out of the boundary of the target object. The negative effect of inaccurate sampling is especially serious for densely located objects. To fix this problem, the learned offsets are further restricted by the boundary of the reference anchor. The modulated offset is calculated by 
\begin{equation}
    \Delta \mathbf{\tilde{p}_q} = \alpha \cdot (w, h) \cdot (\sigma(\Delta\mathbf{p_q}) - 1/2) \cdot R^T(\theta)
\end{equation}
where $\alpha$ is the hyper-parameter controlling the offset range relative to the reference anchor. By default, the $\alpha$ is set to 1.
Based on the modulated offset, the rotation-sensitive deformable attention is calculated as
\begin{equation}
\label{eq:rsdeform}
    \text{RSDeform}(z_q, A_q, x) = \sum_{k=1}^{K}{W_{qk} W^{'} x(\mathbf{p_q} + \Delta \mathbf{\tilde{p}_{qk}})}
\end{equation}
In this way, we can transform the sampling locations to explicitly adapt to arbitrary orientations based on corresponding reference anchors.

The Rotation Sensitive Deformable Attention serves as the basic building block for RotaTR. It is mainly used in two modules. One is the feature alignment module, the other is the decoder layer. The following two parts detail the specific application.

\begin{figure}[t]
\centering 
\subcaptionbox{Deform att}{
\label{}
\includegraphics[width=0.145\textwidth]{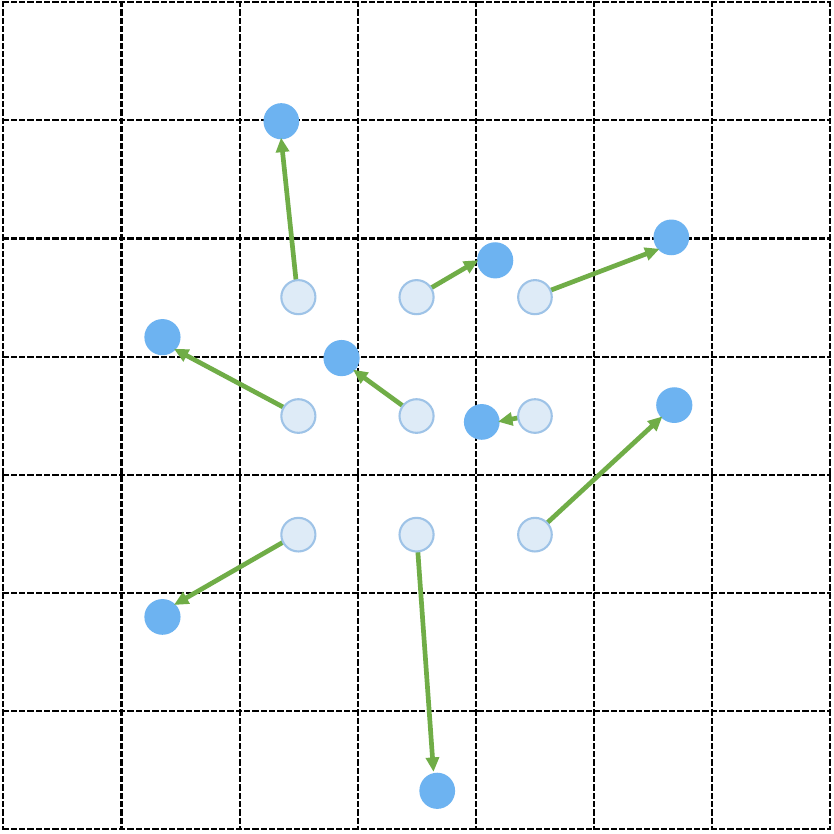}}
\subcaptionbox{Align conv}{
\label{}
\includegraphics[width=0.145\textwidth]{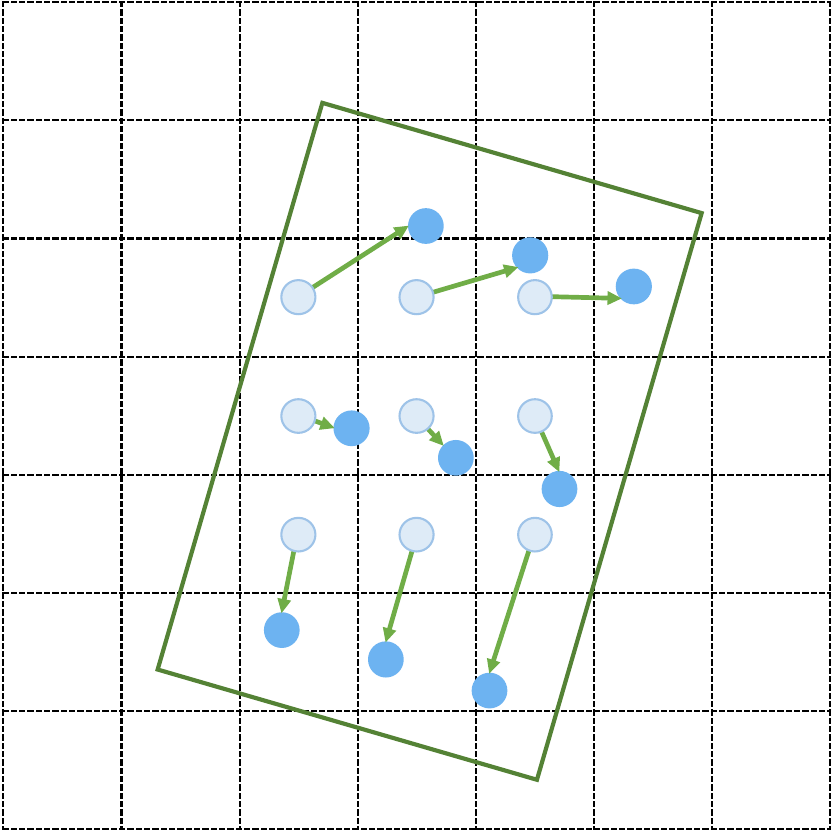}}
\subcaptionbox{RSDeform att}{
\label{}
\includegraphics[width=0.145\textwidth]{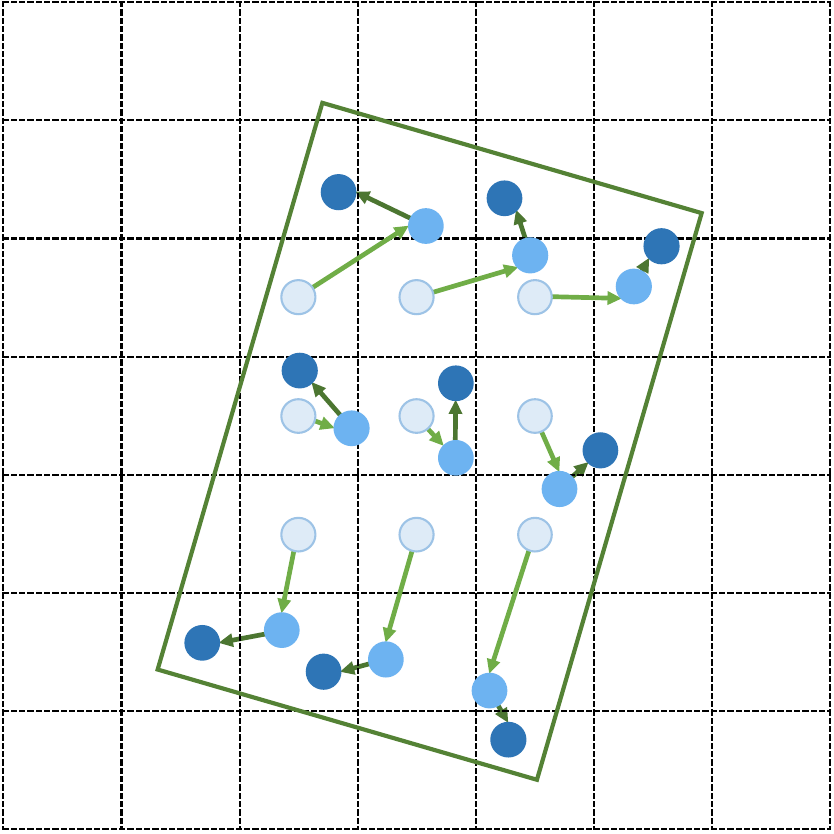}}
\captionsetup{font={small}}
\caption{Comparison between different methods for feature alignment. The light blue points denote the initial locations. The arrows represent the calculated offset vectors. (a) is the original deformable attention, it adaptively learns the offset locations. (b) is the alignment convolution proposed in S2ANet\cite{S2ANet}. It explicitly computes the offset locations by given orientations. (c) is our proposed RSDeformable attention. It can be viewed as a two-step alignment process. The first step is to get the explicitly computed offset locations. The second step is to adaptively learn the offset locations. }
\label{fig:deform_plot}
\vspace{-6pt}
\end{figure}

\textbf{Comparison with Related Operations}
We compare the RSDeform Attention with two most related operators that are commonly used to relieve the feature misalignment problem caused by arbitrary orientation. For clarity, we present the schematic plot in Figure-\ref{fig:deform_plot}.
The first related operator is the original form of deformable attention. It learns an offset field to generate the sampling locations. During training, the offset field is adaptive to fit the target object. The second one is the align convolution proposed in S2ANet\cite{S2ANet}. Align convolution is built upon the standard convolution and explicitly calculates the offset generated by the orientation. In contrast, the RSDeform attention absorbs the advantages of these two. On one hand, it explicitly calculates the offset field like align convolution, on the other hand, it keeps the capability of adaptive fitting like deformable attention. 

\subsection{Feature Alignment Module}
\label{sec:feature_align}
\begin{figure}[t!]
    \centering
    \includegraphics[width=0.8\linewidth]{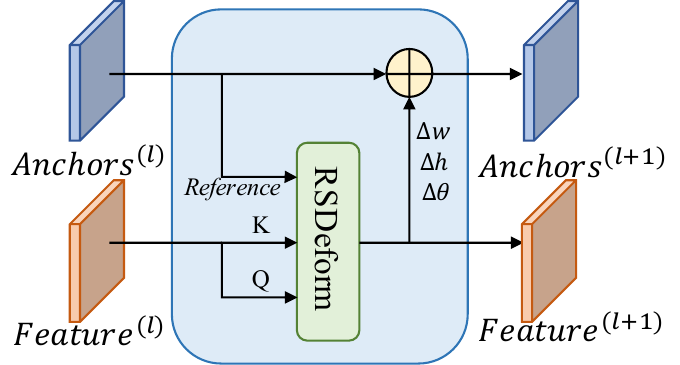}
    \captionsetup{font={small}}
    \caption{Schematic overview of the feature alignment module of the encoder.}
    \label{fig:fl_module}
    \vspace{-6pt}
\end{figure}

% two extra branch
% pipeline: preset anchors, classify, regress, align
% inference notice
Feature alignment module can be viewed as part of the encoder network. It has a similar structure to the encoder layer, which takes the encoded multi-level features as input. 
The difference is that, besides attention layers and feed-forward layers, the feature alignment module has two extra branches: anchor classification branch and anchor regression branch. 
The classification branch distinguishes each pixel of features into foreground and background and the regression branch predicts the extra offset with respect to predefined horizontal anchors for each pixel location. 
Then the generated rotated anchors are fed into the RSDeform attention as the reference anchors, by which the encoded features are further refined.
The whole alignment pipeline is depicted in Figure-\ref{fig:fl_module}.
By default, the anchor classification branch is discarded in the inference phrase to speed up the inference as we only need the rotated anchors to refine the features. 

% \begin{figure}[t!]
%     \centering
%     \includegraphics[width=0.8\linewidth]{pics/fl_module.pdf}
%     \captionsetup{font={small}}
%     \caption{Schematic overview of the feature alignment module of the encoder.}
%     \label{fig:fl_module}
%     \vspace{-6pt}
% \end{figure}

\subsection{Rotation Sensitive Decoder Layer}
% three parts
% inputs, q, k, v
% iterative refinement
% dynamic anchor update
The rotation-sensitive decoder layer is composed of three parts: a standard multi-head self-attention layer, a RSDeform cross-attention layer and a feed-forward layer. 
In the self-attention layer, the query, key and value are defined as Eq-\ref{eq:self}.
% In the self-attention layer, the query, key and value share the same content items while the query and key contains extra positional items. The only difference with the original self attention is that the positional items here is extra encoded by the angular information. Shown in Eq-\ref{eq:pos_q}.
In the cross attention layer, we follow DAB-DETR to put a $\text{MLP}^{(s)}$ to obtain a scalar vector conditional on the content information $c_q$ and use it to perform element-wise multiplication with the positional encoding:
\begin{equation}
    \text{Query}: c_q + p_q\cdot \text{MLP}^{(s)}(c_q)
\end{equation}
Then, the rotation-sensitive cross attention is calculated by Eq-\ref{eq:rsdeform}.

Following the previous practice, we use the decoder layer to predict the relative reference offset ($\delta \hat{x}$, $\delta \hat{y}$, $\delta \hat{w}$, $\delta \hat{h}$, $\delta \hat{\theta}$) and update the reference anchors layer-by-layer.

\subsection{Optimization Objective}
% point set loss instead of L1 loss
% label assignment: alignment assignment. decoder assignment.
The horizontal detectors usually use L1 loss to regress the bounding boxes. It has two problems. The first is that the value range of coordinates differs far away from the angle range. A weighting factor must be introduced to balance the two regression tasks, while finding the proper weighting factor is not an easy task. The second is that each rotated box has two equivalent representations. We can simply exchange the position of width and height and modify the rotation angle correspondingly to get the other representation. This leads to the target ambiguity problem. These two problems have been thoroughly discussed in previous work\cite{Yang2019SCRDetTM}. To fix this, we propose a simple point set loss for regression. For a predicted box and a target box, we first transform them into two point sets, representing each box's eight corner points, $[p_i]$ and $[q_i]$. Note that the point set is in the order of either clockwise or counterclockwise. The point set loss is calculated by
\begin{equation}
    \mathcal{L}_r = \inf_{\pi}\sum_i\|p_i - q_{\pi(i)}\|
    \vspace{-6pt}
\end{equation}
where $\pi(i)$ is an index mapping function from set $\{0,1,2,3\}$ to itself. The point set loss finds the minimum distance sum between the given two point sets. As the two sets are in order, the number of mapping is limited. Thus the point set loss can be calculated quickly. The final optimization objective is the combination of the focal loss, regression loss and the rotated-IoU loss.

\section{Experiment}
\subsection{Datasets}
We evaluate our method on multiple datasets mainly covering aerial images: DOTA-1.0 and DOTA-1.5\cite{xia2018dota}, ship detection HRSC2016\cite{liu2017hrsc}, and text detection MSRA-TD500\cite{yao2012msra_tda}.

\textbf{DOTA}\cite{xia2018dota} is one of the largest datasets for multi-class object detection in aerial images with two released versions: DOTA-v1.0 and DOTA-v1.5. \textbf{DOTA-v1.0} contains 15 categories, 2,806 images and 188,282 instances. Images' scales range from $800 \times 800$ to $4000 \times 4000$ pixels. 
% In the dataset, each object is annotated by an oriented bounding box (OBox), which can be denoted as ($x_1,y_1,x_2,y_2,x_3,y_3/,x_4,y_4$), where ($x_i,y_i$) denotes the i-th vertice of OBox. 
It is split into training, validation, and test sets with 1/2, 1/6, and 1/3 ratios, respectively. For the short names in Table-\ref{tab:dota_comp}, they are defined as (abbreviation-full name): BR-Bridge, SV-Small vehicle, LV-Large vehicle, SH-Ship, HA-Harbor, ST-Storage tank, RA-Roundabout, PL-Plane, BD-Baseball diamond. 
\textbf{DOTA-v1.5} shares identical image sets with v1.0. It is released with a new category, Container Crane (CC). In DOTA-v1.5, more small objects are annotated, making it a much more challenging task compared with v1.0.

For both versions of DOTA datasets, we crop the images into $1024 \times 1024$ patches with a stride of 120. In the single scale setting, we only adopt random horizontal flipping during training to avoid over-fitting and no other tricks are utilized if not specified. In the multi-scale setting, for fair comparison with other methods, we adopt data augmentation (i.e., random rotation) in the training phase and three scales (0.5, 1.0, 1.5) are used to scale the cropped patches.

\textbf{HRSC2016}\cite{liu2017hrsc} is another challenging aerial images dataset that contains lots of large aspect ratio ship instances with arbitrary orientation.
% It contains images from two scenarios including ships on sea and ships close inshore. HRSC2016 dataset has four ship categories, 1061 images and 2976 samples in total. 
It is split into training, validation and test set which contains 436 images including 1207 samples, 181 images including 541 samples and 444 images respectively. We use both the training and validation sets for training and test sets for testing. 

\textbf{MSRA-TD500}\cite{yao2012msra_tda} is commonly used for oriented scene text detection and spotting. It contains 300 training images and 200 testing images. Extra 400 images from HUST\cite{yao201hust} are also included for training.
% We perform experiments on these scene text detection datasets in order to verify the scenario generality of our work.

\subsection{Implementation Details}
We implement the proposed method on MMDetection\cite{mmdetection}. In all experiments, we adopt Deformable DETR-O with ResNet-50 backbone (pretrained on ImageNet) as the baseline method. It is the modified version of Deformable DETR to oriented detection. We use multi-scale features from C3 to C5 of ResNet-50. 
We train the network with AdamW for 50 epochs. 
In the first 40 epochs, the learning rate is $1e-4$ and then $1e-5$ for another 10 epochs. The momentum and weight decay are set to 0.9 and 0.001 respectively. We train our method on a single A100 GPU with the batch size of 4. The loss weights for classification, regression and rotated-IoU are set as 2, 5 and 2, respectively. In the inference stage, we follow the same scale setting as training. No post-processing is needed for associating objects.

\begin{table*}[ht!]
\begin{adjustbox}{max width=\textwidth}
\begin{tabular}{c|c|c|ccccccccccccccc|c}
\hline
Model                        & Backbone  & Epochs & PL                           & BD                           & $\text{BR}^{\dagger}$        & GTF                          & $\textbf{SV}^{\dagger}$      & $\textbf{LV}^{\dagger}$      & $\text{SH}^{\dagger}$        & TC                           & BC                           & ST                           & SBF                          & RA                           & ${HA}^{\dagger}$             & SP                           & HC                           & mAP                          \\ \hline
$Single Scale$               &           &        &                              &                              &                              &                              &                              &                              &                              &                              &                              &                              &                              &                              &                              &                              &                              &                              \\ \hline
Retinanet-O\cite{2017Focal}                 & R-50-FPN  & 12     & 88.67                        & 77.62                        & 41.81                        & 58.17                        & 74.58                        & 71.64                        & 79.11                        & 90.29                        & 82.18                        & 74.32                        & 54.75                        & 60.60                        & 62.57                        & 69.67                        & 60.64                        & 68.43                        \\
$\text{Retinanet-O}^{*}$     & R-50-FPN  & 36     & 88.83                        & 71.62                        & 43.42                        & 63.91                        & 68.22                        & 73.61                        & 86.65                        & 90.82                        & 86.17                        & 83.79                        & 55.82                        & 65.28                        & 65.83                        & 67.66                        & 58.71                        & 71.72                        \\
                             & R-50-FPN  & 50     & 88.62                        & 71.21                        & 41.70                        & 64.36                        & 68.08                        & 73.50                        & 86.85                        & 90.27                        & 85.47                        & 83.51                        & 55.31                        & 62.78                        & 65.41                        & 68.45                        & 58.20                        & 71.12                        \\
RSDet\cite{Qian2022RSDetPM}                        & R-101-FPN & 12     & 89.80                        & 82.90                        & 48.60                        & 65.20                        & 69.50                        & 70.10                        & 70.20                        & 90.50                        & 85.60                        & 83.40                        & {\color[HTML]{3531FF} 62.50} & 63.90                        & 65.60                        & 67.20                        & {\color[HTML]{3531FF} 68.00} & 72.20                        \\
CenterMap\cite{Wang2021LearningCP}                    & R-50-FPN  & 12     & 88.88                        & 81.24                        & 53.15                        & 60.65                        & 78.62                        & 66.55                        & 78.10                        & 88.83                        & 77.80                        & 83.61                        & 49.36                        & 66.19                        & 72.10                        & 72.36                        & 58.70                        & 71.74                        \\
R3Det\cite{R3Det}                        & R-101-FPN & 12     & 88.76                        & 83.09                        & 50.91                        & 67.27                        & 76.23                        & 80.39                        & 86.72                        & 90.78                        & 84.68                        & 83.24                        & 61.98                        & 61.35                        & 66.91                        & 70.63                        & 53.94                        & 73.79                        \\
SCRDet\cite{Yang2019SCRDetTM}                       & R-101-FPN & 12     & {\color[HTML]{FE0000} 89.98} & 80.65                        & 52.09                        & 68.36                        & 68.36                        & 60.32                        & 72.41                        & 90.85                        & {\color[HTML]{FE0000} 87.94} & {\color[HTML]{FE0000} 86.86} & {\color[HTML]{FE0000} 65.02} & 66.68                        & 66.25                        & 68.24                        & 65.21                        & 72.61                        \\
S2ANet\cite{S2ANet}                       & R-50-FPN  & 12     & {\color[HTML]{333333} 89.11} & 82.84                        & 48.37                        & 71.11                        & 78.11                        & 78.39                        & 87.25                        & 90.83                        & 84.90                        & 85.64                        & 60.36                        & 62.60                        & 65.26                        & 69.13                        & 57.94                        & 74.12                        \\
$\text{S2ANet}^{*}$          & R-50-FPN  & 36     & 88.57                        & 83.61                        & {\color[HTML]{3531FF} 53.21} & 71.32                        & 77.08                        & 81.69                        & 87.35                        & 90.27                        & 85.91                        & 85.14                        & 52.27                        & 65.09                        & 73.70                        & 66.37                        & 65.80                        & 75.21                        \\
                             & R-50-FPN  & 50     & 88.62                        & 81.63                        & 52.91                        & 71.28                        & {\color[HTML]{FE0000} 79.08} & 80.22                        & {\color[HTML]{FE0000} 88.21} & 90.88                        & 84.24                        & 83.24                        & 59.20                        & 65.12                        & 76.11                        & 71.85                        & 48.03                        & 74.71                        \\
$\text{RoI-Transformer}^{*}$\cite{ding2019roitrans} & R-50-FPN  & 36     & 88.96                        & 81.30                        & 47.89                        & 65.44                        & 80.71                        & 81.62                        & 87.76                        & 90.90                        & 83.90                        & 85.43                        & 58.10                        & 65.06                        & 72.66                        & {\color[HTML]{3531FF} 72.97} & 44.68                        & 73.83                        \\
$\text{Oriented-RCNN}^{*}$\cite{orcnn}   & R-50-FPN  & 12     & {\color[HTML]{3531FF} 89.46} & 82.12                        & 54.78                        & 70.86                        & 78.93                        & 83.00                        & 88.20                        & {\color[HTML]{FE0000} 90.90} & {\color[HTML]{3531FF} 87.50} & 84.68                        & 63.97                        & {\color[HTML]{FE0000} 67.69} & 74.94                        & 68.84                        & 52.28                        & 75.60                        \\
                             & R-50-FPN  & 36     & 88.90                        & 82.91                        & {\color[HTML]{FE0000} 55.11} & 71.14                        & 78.99                        & {\color[HTML]{3531FF} 83.06} & 88.14                        & {\color[HTML]{3531FF} 90.90} & 83.77                        & 85.49                        & 61.76                        & {\color[HTML]{3531FF} 66.76} & 76.62                        & 68.85                        & 53.41                        & {\color[HTML]{3531FF} 75.72} \\
                             & R-50-FPN  & 50     & 88.65                        & {\color[HTML]{FE0000} 83.42} & 52.95                        & {\color[HTML]{FE0000} 73.27} & 78.88                        & 83.77                        & 87.99                        & 90.88                        & 79.69                        & 85.61                        & 56.67                        & 62.80                        & {\color[HTML]{FE0000} 77.62} & 69.65                        & 52.17                        & 74.94                        \\
O2Transformer\cite{o2detr}                & R-50      & 50     & 83.89                        & 75.11                        & 44.04                        & 64.20                        & 78.39                        & 76.78                        & 87.68                        & 90.60                        & 78.58                        & 71.82                        & 53.21                        & 60.35                        & 55.36                        & 61.90                        & 47.89                        & 68.65                        \\
FT-O2Transformer             & R-50-FPN  & 50     & 88.76                        & 81.91                        & 51.20                        & {\color[HTML]{3531FF} 72.18} & 77.64                        & 80.47                        & 87.84                        & 90.85                        & 84.56                        & 81.68                        & 61.42                        & 64.61                        & 67.50                        & 64.28                        & 62.15                        & 74.47                        \\
Deform-DETR-O\cite{deformdetr}                & R-50      & 50     & 86.91                        & 74.40                        & 47.22                        & 64.31                        & 72.07                        & 76.01                        & 86.66                        & 90.84                        & 78.45                        & 77.57                        & 53.28                        & 60.50                        & 62.41                        & 68.16                        & 50.47                        & 69.95                        \\
RotaTR                     & R-50      & 50     & 88.87                        & {\color[HTML]{3531FF} 82.91} & 49.67                        & 69.52                        & {\color[HTML]{3531FF} 79.01} & {\color[HTML]{FE0000} 83.58} & {\color[HTML]{3531FF} 88.20} & 90.64                        & 78.99                        & {\color[HTML]{3531FF} 85.78} & 54.85                        & 63.44                        & {\color[HTML]{3531FF} 77.05} & {\color[HTML]{FE0000} 73.12} & {\color[HTML]{FE0000} 69.06} & {\color[HTML]{FE0000} 75.86} \\ \hline
$Multi-Scale$                &           &        &                              &                              &                              &                              &                              &                              &                              &                              &                              &                              &                              &                              &                              &                              &                              &                              \\ \hline
$\text{Retinanet-O}^{*}$\cite{2017Focal}     & R-50-FPN  & 36     & {\color[HTML]{FE0000} 89.01} & 84.13                        & 50.92                        & 70.33                        & 71.08                        & 77.21                        & 84.33                        & 90.80                        & 71.63                        & 86.42                        & 61.78                        & 64.25                        & 67.98                        & 69.20                        & 48.04                        & 76.50                        \\
$\text{RoI-Transformer}^{*}$\cite{ding2019roitrans} & R-101-FPN & 36     & 88.51                        & {\color[HTML]{FE0000} 86.43} & {\color[HTML]{FE0000} 60.74} & {\color[HTML]{FE0000} 80.12} & 77.47                        & 85.10                        & 88.49                        & {\color[HTML]{3531FF} 90.91} & {\color[HTML]{FE0000} 88.67} & 84.73                        & {\color[HTML]{3531FF} 70.42} & 65.78                        & 79.12                        & {\color[HTML]{3531FF} 80.94} & 74.65                        & {\color[HTML]{3531FF} 80.11} \\
O2DETR\cite{o2detr}                       & R-50      & 50     & 86.01                        & 75.92                        & 46.02                        & 66.65                        & 79.70                        & 79.93                        & 89.17                        & 90.44                        & 81.19                        & 76.00                        & 56.91                        & 62.45                        & 64.22                        & 65.80                        & 58.96                        & 72.15                        \\
FT-O2DETR                    & R-50-FPN  & 50     & {\color[HTML]{3531FF} 88.89} & 83.41                        & 56.72                        & 79.75                        & {\color[HTML]{3531FF} 79.89} & {\color[HTML]{3531FF} 85.45} & {\color[HTML]{3531FF} 89.77} & 90.84                        & 86.15                        & {\color[HTML]{FE0000} 87.66} & 69.94                        & {\color[HTML]{FE0000} 68.97} & 78.83                        & 78.19                        & 70.38                        & 79.66                        \\
RotaTR                     & R-50      & 50     & 88.50                        & 85.91                        & 60.13                        & {\color[HTML]{3531FF} 78.94} & 79.80                        & 84.95                        & {\color[HTML]{FE0000} 89.88} & {\color[HTML]{FE0000} 90.94} & {\color[HTML]{3531FF} 88.47} & 86.43                        & 68.58                        & 67.31                        & {\color[HTML]{FE0000} 81.59} & 76.88                        & {\color[HTML]{3531FF} 76.93} & 79.95                        \\
                             & R-101     & 50     & 88.31                        & {\color[HTML]{3531FF} 85.64} & {\color[HTML]{3531FF} 60.44} & 78.81                        & {\color[HTML]{FE0000} 79.91} & {\color[HTML]{FE0000} 86.22} & 89.34                        & 90.90                        & 87.71                        & {\color[HTML]{3531FF} 86.44} & {\color[HTML]{FE0000} 71.05} & {\color[HTML]{3531FF} 68.67} & {\color[HTML]{3531FF} 79.20} & {\color[HTML]{FE0000} 82.31} & {\color[HTML]{FE0000} 77.56} & {\color[HTML]{FE0000} 80.48} \\ \hline
\end{tabular}
\end{adjustbox}
\captionsetup{font={small}}
\caption{Comparison to state-of-art on DOTA-v1.0 test set. Results for each class are reported.  The best and second best results are colored in \textcolor{red}{red} and \textcolor{blue}{blue} respectively. 
Classes with $\dagger$ superscript denotes the large aspect ratio targets.
$\textbf{SV}$ and $\textbf{LV}$ are in bold. They are most likely to be dense located.
The method with $*$ superscript denotes that the reproduced results. Most of the reproduced results surpass that in the original work.}
\label{tab:dota_comp}
\vspace{-10pt}
\end{table*}

\subsection{Comparison Results}
\textbf{Competitors.} On DOTA dataset and HRSC dataset, we mainly compare our method to several CNN-based methods (i.e., Retinanet-O, RSDet\cite{Qian2022RSDetPM}, CenterMap\cite{Wang2021LearningCP}, R3Det\cite{R3Det}, SCRDet\cite{Yang2019SCRDetTM}, S2ANet\cite{S2ANet}) and transformer-based methods (i.e., RoI-Transformer\cite{ding2019roitrans}, O2Transformer\cite{o2detr}, Deformable-DETR-O). The Retinanet-O and Deformable-DETR-O are the modified version of the original horizontal Retinanet\cite{2017Focal} and Deformable-DETR\cite{deformdetr} to oriented detection, respectively.
The majority of them are CNN-based methods. Only RoI Transformer and O2DETR are related to transformer. Note that the RoI transformer is not an end-to-end transformer-based detector. It uses transformer only for region proposal generation and should also be categorized into CNN-based detectors. On MSRA-TD500 datasets, we choose the commonly used text detectors as the competitors. 

\textbf{Training Length.} On DOTA dataset, most of the CNN-based detectors adopt the 12-epoch training setting. Actually, increasing the training length to 36 epochs will greatly increase the performance. Continually increasing the training length to 50 won't improve the performance but will overfit. While for our transformer-based methods, 12-epoch training is not enough for convergence. Defiantly, we train 50 epochs for comparison. The training lengths on other datasets are also adjusted according to the actual convergence state.

\textbf{Results on DOTA.} 
The comparison results on DOTA-v1.0 is shown in Table-\ref{tab:dota_comp}.
We can see that the transformer-based methods perform much worse than the modern CNN-based methods.
For example, the newly proposed O2Transformer gets mAP of 68.65, even inferior to the Retinanet-O (71.72).
The Deformable DETR-O gets 69.95. It is also at the same level as O2DETR. 
By contrast, the proposed RotaTR achieve 75.86\% mAP with only ResNet 50 backbone in the single scale, outperforming all the competitors. 
Note that RotaTR gets very challenging results on SV (small vehicle) and LV (large vehicle). These two are most likely to be densely located.
% In multi-scale experiments, our RotaTR achieves 79.95\% and 80.40\% mAP with ResNet-50 and ResNet-101 backbone, respectively. 

On DOTA-v1.5 dataset, we report the AP metric under 50\% and 75\% IoU conditions. The results are shown in Table-\ref{tab:dota_1_1_5}. Specifically, RotaTR gets 70.5\% mAP under IoU 50\% and 46.2\% mAP under IoU 75\%,  greatly surpassing the baseline method by 4.7\% and 10.7\%, respectively. 

\begin{table}[]
\begin{adjustbox}{max width=\linewidth}
\begin{tabular}{c|ccc|ccc}
\hline
\multirow{2}{*}{Method} & \multicolumn{3}{c|}{v1.0 train/val} & \multicolumn{3}{c}{v1.5 train/val} \\ \cline{2-7} 
                        & AP50      & AP75      & AP50:95     & AP50      & AP75     & AP50:95     \\ \hline
Retinanet-O\cite{2017Focal}               & 66.3      & 41.7      & 38.3        & 67.5      & 41.3     & 40.7        \\
R3Det\cite{R3Det}                   & 67.2      & 38.4      & 38.4        & 69.2      & 44.6     & 42.6        \\
RoI Transformer\cite{ding2019roitrans}         & 71.3      & 47.5      & 44.8        & 70.2      & 46.1     & 43.5        \\
O2DETR\cite{o2detr}                  & 65.3      & 41.4      & 39.8        & 63.8      & 33.2     & 35.1        \\
Deformable-DETR-O\cite{deformdetr}        & 67.3      & 43.7      & 40.1        & 65.8      & 35.5     & 36.53       \\
RotaTR                & \textbf{72.0}      & \textbf{49.0}      & \textbf{45.1}        & \textbf{70.5}      & \textbf{46.2}     & \textbf{43.6}        \\ \hline
\end{tabular}
\end{adjustbox}
\captionsetup{font={small}}
\caption{Comparison on DOTA-v1.0 and v1.5 datasets}
\label{tab:dota_1_1_5}
\vspace{-12pt}
\end{table}

\textbf{Results on HRSC.} The HRSC2016 dataset contains only one class and the training set has only 617 images. We find that the original training setting (e.g., 50 epochs) is hard for the RotaTR to fully converge. Thus, for this dataset, the training steps are extended to five times of the original. The results are shown in Table-\ref{tab:hrsc}. It is seen that RotaTR outperforms the baseline method by 2\% mAP and achieves comparable performance to the state-of-the-art. 

\begin{table}[b]
\centering
\begin{adjustbox}{max width=\linewidth}
\begin{tabular}{c|c|cc}
\hline
Methods         & Backbone  & mAP (07)       & mAP (12)       \\ \hline
DRN\cite{pan2020dynamic}             & H-34      & -             & 92.7          \\
R3Det\cite{R3Det}         & R-101-FPN & 89.26         & 96.01         \\
S2A-Net\cite{S2ANet}         & R-101-FPN & 90.17         & 95.01         \\
RRPN\cite{ma2018arbitrary}     & R-101     & 79.08         & 85.64         \\
R2CNN\cite{jiang2017r2cnn}           & R-101     & 73.07         & 79.73         \\
Oriented R-CNN\cite{orcnn}  & R-50-FPN  & \textbf{90.4} & 96.5          \\
RoI Transformer\cite{ding2019roitrans} & R-101-FPN & 86.2          & -             \\
O2Transformer\cite{o2detr}   & R-101-FPN & 88.2          & 93.3          \\
Deformable DETR-O\cite{deformdetr} & R-50      & 88.4          & 94.1          \\
RotaTR        & R-50      & 90.3          & \textbf{96.7} \\ \hline
\end{tabular}
\end{adjustbox}
\captionsetup{font={small}}
\caption{Results on HRSC2016}
\vspace{-6pt}
\label{tab:hrsc}
\end{table}

\textbf{Results on MSRA-TD500.} The comparison in MSRA-TD500 is shown in Table-\ref{tab:msra_comp}. Similar to HRSC, the training epoch is modified to 250 for fully converge. It is seen that RotaTR also gets very challenging results in scene text detection.

\begin{table}[]
\centering
% \begin{adjustbox}{max width=\linewidth}
\setlength{\tabcolsep}{3.5mm}
\begin{tabular}{c|ccc}
\hline
Method   & Precision     & Recall        & H-mean        \\ \hline
RRPN\cite{ma2018arbitrary}     & 82.0          & 68.0          & 74.0          \\
DeepReg\cite{DeepReg}  & 77.0          & 70.0          & 74.0          \\
EAST\cite{EAST}     & 87.3 & 67.4          & 76.1          \\
SegLink\cite{shi2017detecting}  & 86.0          & 70.0          & 77.0          \\
PAN\cite{PAN}      & 80.7          & 77.3 & 78.9          \\
RRD\cite{RRD}      & 87.0          & 73.0          & 79.0          \\
MSR\cite{Xue2019MSRMS}       & 87.4          & 76.7          & 81.7          \\
SAE\cite{Tian_2019_CVPR}       & 84.2          & \textbf{81.7} & 82.9          \\
RotaTR & \textbf{90.9}          & 77.6         & \textbf{83.8} \\ \hline
\end{tabular}
% \end{adjustbox}
\captionsetup{font={small}}
\caption{Results on MSRA-TD500}
\label{tab:msra_comp}
\vspace{-12pt}
\end{table}

\begin{figure*}[t!]
\centering  
\subcaptionbox{ours}{
\includegraphics[width=0.225\textwidth]{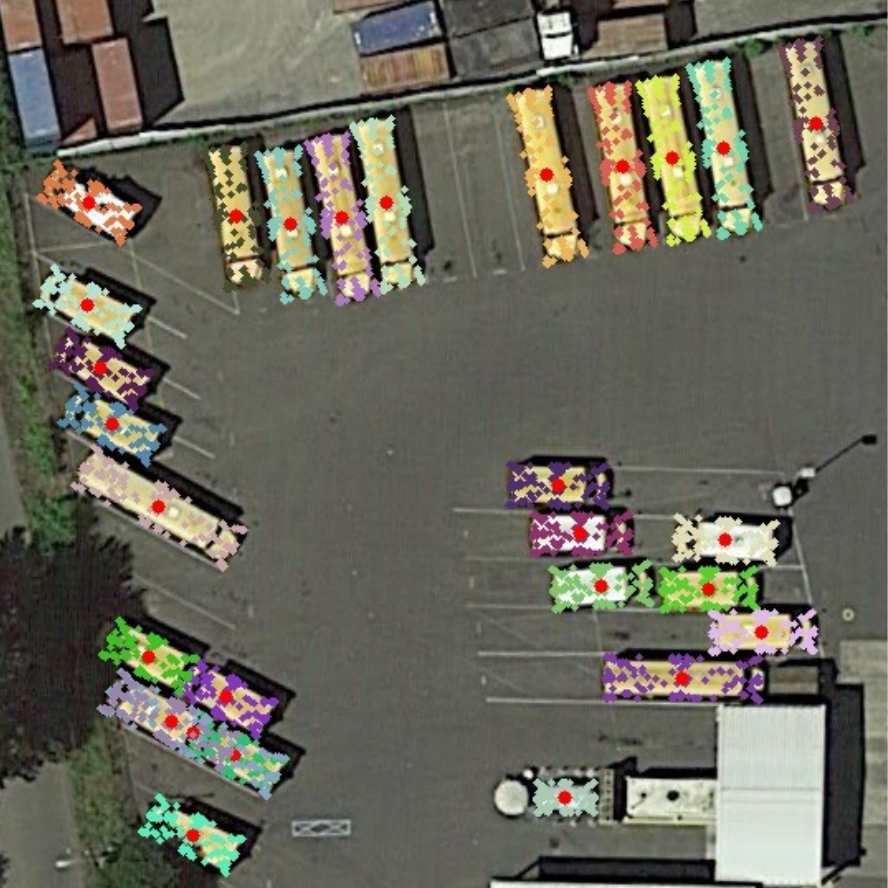}}
\subcaptionbox{no range restriction}{
\includegraphics[width=0.225\textwidth]{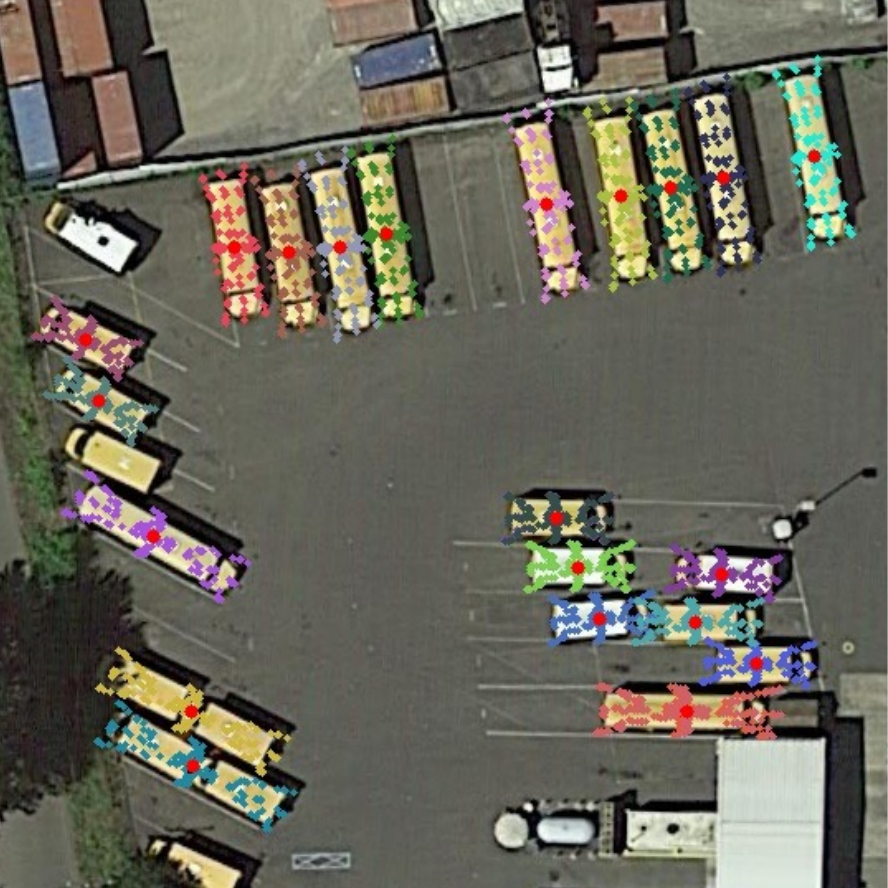}}
\subcaptionbox{no anchor shape modulation}{
\includegraphics[width=0.225\textwidth]{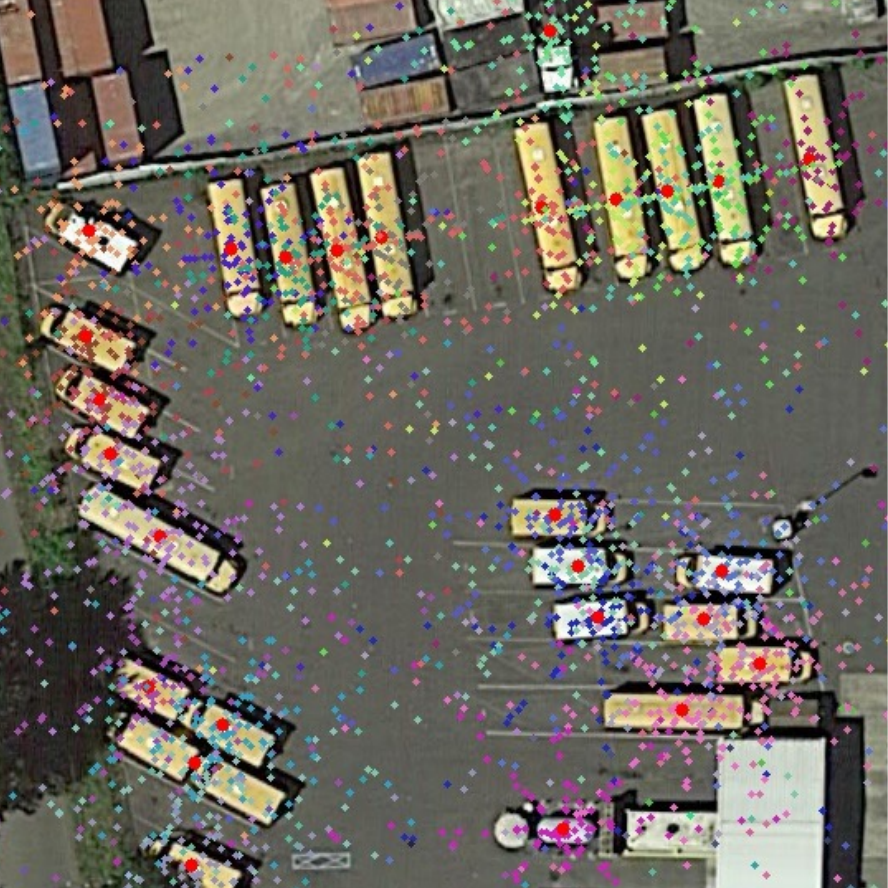}}
\subcaptionbox{no orientation modulation}{
\includegraphics[width=0.225\textwidth]{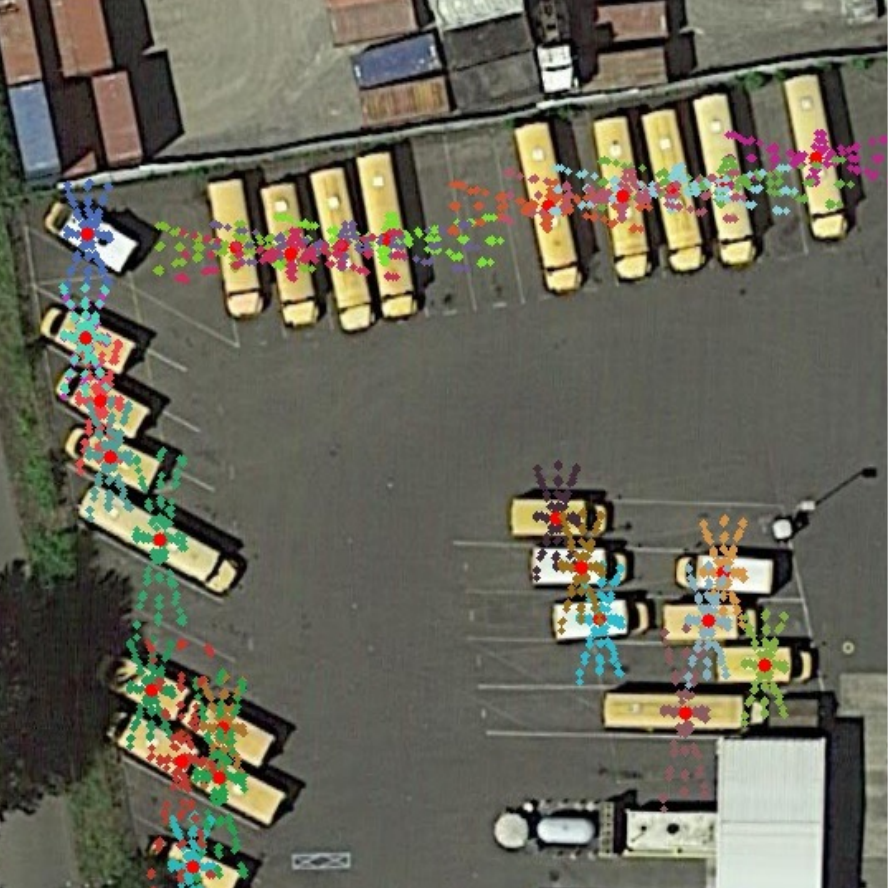}}
\captionsetup{font={small}}
\caption{Sampled points of the deformable cross attention under different configurations. Red points represent the detected instance center, other colored points represent the sampled points for each instance.
(a) Our default configuration. The sampled points are well aligned to the targets' geometry. (b) Dropping the sampled points' range restriction. Some points get out of the range of the objects. (c) Dropping the modulation from width and height. The sampled points span the whole image. (d) Dropping the modulation from the orientation. It represents the original Deformable Attention. The sampled points learn the width and height well, but the orientation is totally incorrect.}
\label{fig:pts_plot}
\vspace{-12pt}
\end{figure*}

\subsection{Ablation Study}
By default, the DOTA-v1.0 validation set is used for ablation study.

\begin{table}[b]
\setlength{\tabcolsep}{4mm}
\centering
\begin{tabular}{cccc|l}
\hline
DAB          & RS           & FA           & PS           & AP50       \\ \hline
             &              &              &              & 65.4       \\
$\checkmark$ &              &              &              & 66.0 (\textbf{+0.6}) \\
             & $\checkmark$ &              &              & 69.3 (\textbf{+3.9}) \\
$\checkmark$ & $\checkmark$ &              &              & 69.4 (\textbf{+4.0}) \\
$\checkmark$ &              & $\checkmark$ &              & 66.2 (\textbf{+0.8}) \\
$\checkmark$ & $\checkmark$ & $\checkmark$ &              & 70.8 (\textbf{+5.4}) \\
$\checkmark$ &              &              & $\checkmark$ & 67.2 (\textbf{+1.8}) \\
$\checkmark$ & $\checkmark$ & $\checkmark$ & $\checkmark$ & 72.0 (\textbf{+6.6}) \\ \hline
\end{tabular}
\captionsetup{font={small}}
\caption{Ablation study of proposed modules. DOTA-v1.0 is used in this experiment. The results are reported on the DOTA-v1.0 validation set. \textbf{DAB} means the dynamic anchor box mechanism. \textbf{RS} means the rotation-sensitive cross attention. \textbf{FA} means the feature alignment module. \textbf{PS} means the point set loss.}
\label{tab:modules}
\end{table}

\textbf{The contribution of each part. }
We experiment on DOTA-v1.0 to validate the effectiveness of each proposed module. The results are shown in Table-\ref{tab:modules}. It is seen that the naive Deformable DETR gets only 65.4\% mAP. The DAB mechanism can slightly increase the performance to 66.0. While adding point set loss can increase the performance to 67.2\%. Extra experiments on other detectors shown in Table-\ref{tab:point_set} also show the effectiveness of the point set loss. The largest contribution comes from the RSDeform attention, which boosts the performance by 4\% mAP. We also find that simply adding the feature alignment module brings little improvement. 
The effect of the feature alignment module emerges only when the rotation-sensitive cross-attention module is applied. When all modules are applied, the performance is increased to 72.0\%. 
The effect of the proposed RSDeform can also be proved via the sampled points. As shown in Figure-\ref{fig:pts_plot}, the sampled points from RSDeform can be well aligned with the target boundary. Dropping each property of RSDeform will result in the misalignment. 

% \begin{table}[h]
% \setlength{\tabcolsep}{4mm}
% \centering
% \begin{tabular}{cccc|l}
% \hline
% DAB          & RS           & FA           & PS           & AP50       \\ \hline
%              &              &              &              & 65.4       \\
% $\checkmark$ &              &              &              & 66.0 (\textbf{+0.6}) \\
%              & $\checkmark$ &              &              & 69.3 (\textbf{+3.9}) \\
% $\checkmark$ & $\checkmark$ &              &              & 69.4 (\textbf{+4.0}) \\
% $\checkmark$ &              & $\checkmark$ &              & 66.2 (\textbf{+0.8}) \\
% $\checkmark$ & $\checkmark$ & $\checkmark$ &              & 70.8 (\textbf{+5.4}) \\
% $\checkmark$ &              &              & $\checkmark$ & 67.2 (\textbf{+1.8}) \\
% $\checkmark$ & $\checkmark$ & $\checkmark$ & $\checkmark$ & 72.0 (\textbf{+6.6}) \\ \hline
% \end{tabular}
% \captionsetup{font={small}}
% \caption{Ablation study of proposed modules. DOTA-v1.0 is used in this experiment. The results are reported on the DOTA-v1.0 validation set. \textbf{DAB} means the dynamic anchor box mechanism. \textbf{RS} means the rotation sensitive cross attention. \textbf{FA} means the feature alignment module. \textbf{PS} means the point set loss.}
% \label{tab:modules}
% \end{table}

\begin{table}[]
\centering
\begin{adjustbox}{max width=\linewidth}
\begin{tabular}{c|ccc}
\hline
Methods           & Point Set Loss & AP50          & AP75          \\ \hline
Retinanet\cite{2017Focal}         &                & 66.3          & 41.7          \\
                  & $\checkmark$   & \textbf{67.1} & \textbf{42.5} \\
S2A-Net\cite{S2ANet}           &                & 69.7          & 42.1          \\
                  & $\checkmark$   & \textbf{70.3} & \textbf{42.6} \\
Deformable DETR-O\cite{deformdetr} &                & 66.0          & 41.8          \\
                  & $\checkmark$   & \textbf{67.2} & \textbf{43.7} \\
RotaTR          &                & 70.8          & 48.0          \\
                  & $\checkmark$   & \textbf{72.0} & \textbf{49.0} \\ \hline
\end{tabular}
\end{adjustbox}
\captionsetup{font={small}}
\caption{Effect of Point Set Loss on other detectors. The results are reported on DOTA-v1.0 validation set. }
\label{tab:point_set}
\vspace{-12pt}
\end{table}

\textbf{Multi-level feature maps.} The multi-level feature pyramid is a common technique to improve performance on small targets. 
To verify the influence of the selection of feature maps on the model performance, we experiment on several detectors and report the results on Table-\ref{tab:ms}. It is noticed that although adding the $4\times$ feature map brings considerable performance gain for Retinanet-O (+0.8\%mAP), O2DETR (+2.5\%mAP) and Deformable-DETR-O (+1.4\%mAP), it has little effect for RotaTR. 
We argue that the performance gain mainly comes from small objects, which are more likely to be suppressed by the feature misalignment problem caused by arbitrary orientation. To figure it out, we re-evaluate Deformable-DETR-O and RotaTR on objects smaller than $16 \times 16$. For the former, the APs with C2 and without are 30.7 and 24.3. For RotaTR, the APs are 31.1 and 30.9. 
% So RotaTR explicitly augments the orientation information to improve this issue, making it less dependent on the high resolution feature map.
% Considering the huge computational cost brings by the low level feature map, we only use the feature maps composed by C2,C3,C4 and C5 as input to the transformer encoder.

\begin{table}[h]
\begin{adjustbox}{max width=\linewidth}
\begin{tabular}{c|ccccccc}
\hline
methods       & \multicolumn{5}{c}{MS downsample ratios}                                 & Params & mAP           \\ \cline{2-6}
              & 64           & 32           & 16           & 8            & 4            &        &               \\ \hline
Retinanet-O\cite{2017Focal}   &              &              &              &              &              & 34M    & 60.3          \\
              & $\checkmark$ & $\checkmark$ & $\checkmark$ & $\checkmark$ &              & 37M    & 66.3          \\
              & $\checkmark$ & $\checkmark$ & $\checkmark$ & $\checkmark$ & $\checkmark$ & 38M    & \textbf{67.1} \\ \hline
O2DETR\cite{o2detr}        &              &              &              &              &              & 38M    & 62.2          \\
              & $\checkmark$ & $\checkmark$ & $\checkmark$ & $\checkmark$ &              & 41M    & 66.1          \\
              & $\checkmark$ & $\checkmark$ & $\checkmark$ & $\checkmark$ & $\checkmark$ & 42M    & \textbf{68.6} \\ \hline
Deform-DETR-O\cite{deformdetr} &              &              &              &              &              & 38M    & 62.9          \\
              & $\checkmark$ & $\checkmark$ & $\checkmark$ & $\checkmark$ &              & 39M    & 67.2          \\
              & $\checkmark$ & $\checkmark$ & $\checkmark$ & $\checkmark$ & $\checkmark$ & 40M    & \textbf{68.6} \\ \hline
RotaTR       &              &              &              &              &              & 38M    & 63.1          \\
              & $\checkmark$ & $\checkmark$ & $\checkmark$ & $\checkmark$ &              & 39M    & 72.0          \\
              & $\checkmark$ & $\checkmark$ & $\checkmark$ & $\checkmark$ & $\checkmark$ & 40M    & \textbf{72.3} \\ \hline
\end{tabular}
\end{adjustbox}
\captionsetup{font={small}}
\caption{Ablation on Multi-Level Features. The performance are reported on DOTA-v1.0 validation set.}
\label{tab:ms}
\vspace{-10pt}
\end{table}

\begin{figure}[t]
\centering 
\subcaptionbox{}{
\label{fig:offset_bar}
\includegraphics[width=0.225\textwidth]{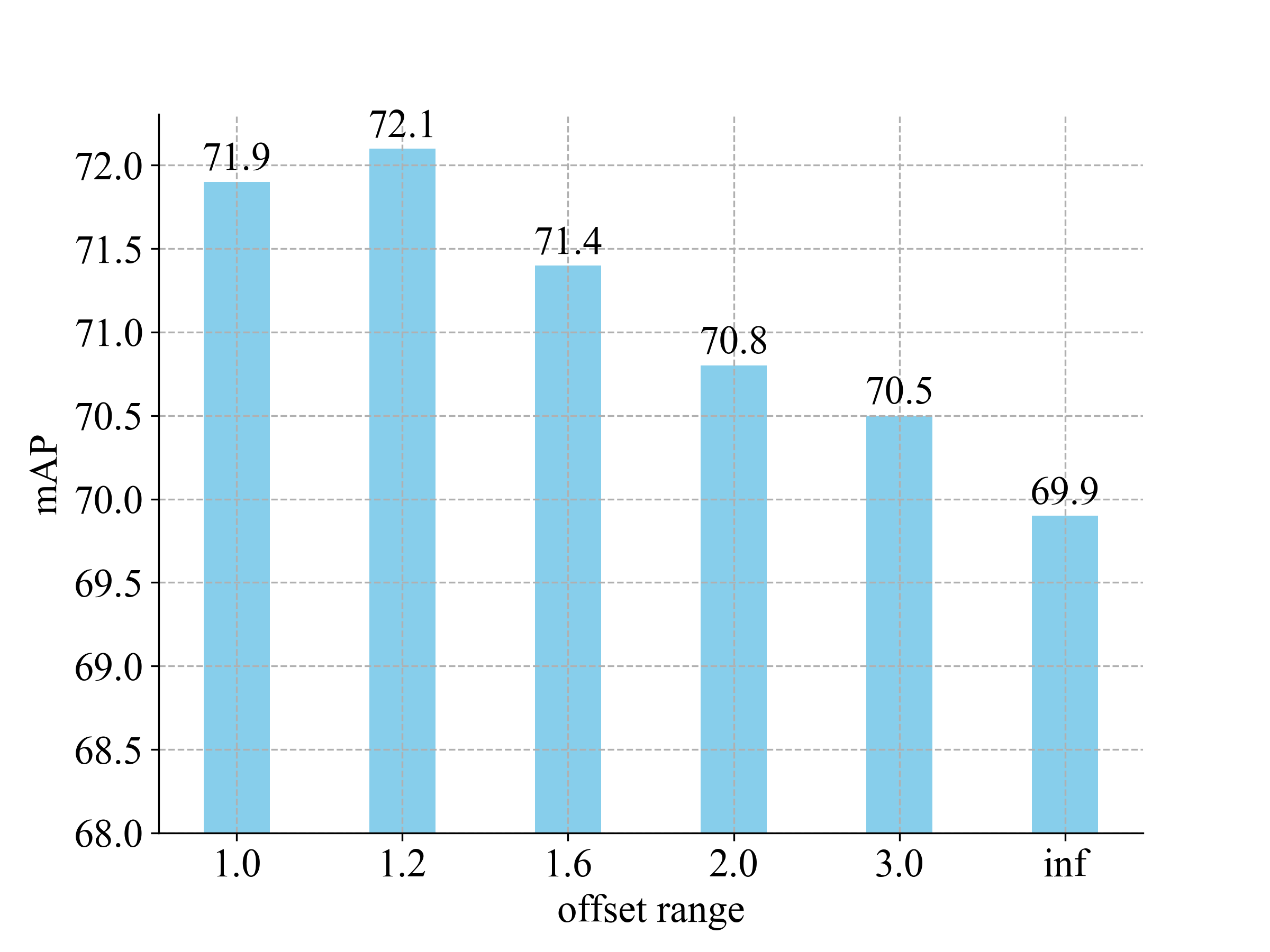}}
\subcaptionbox{}{
\label{fig:nq}
\includegraphics[width=0.225\textwidth]{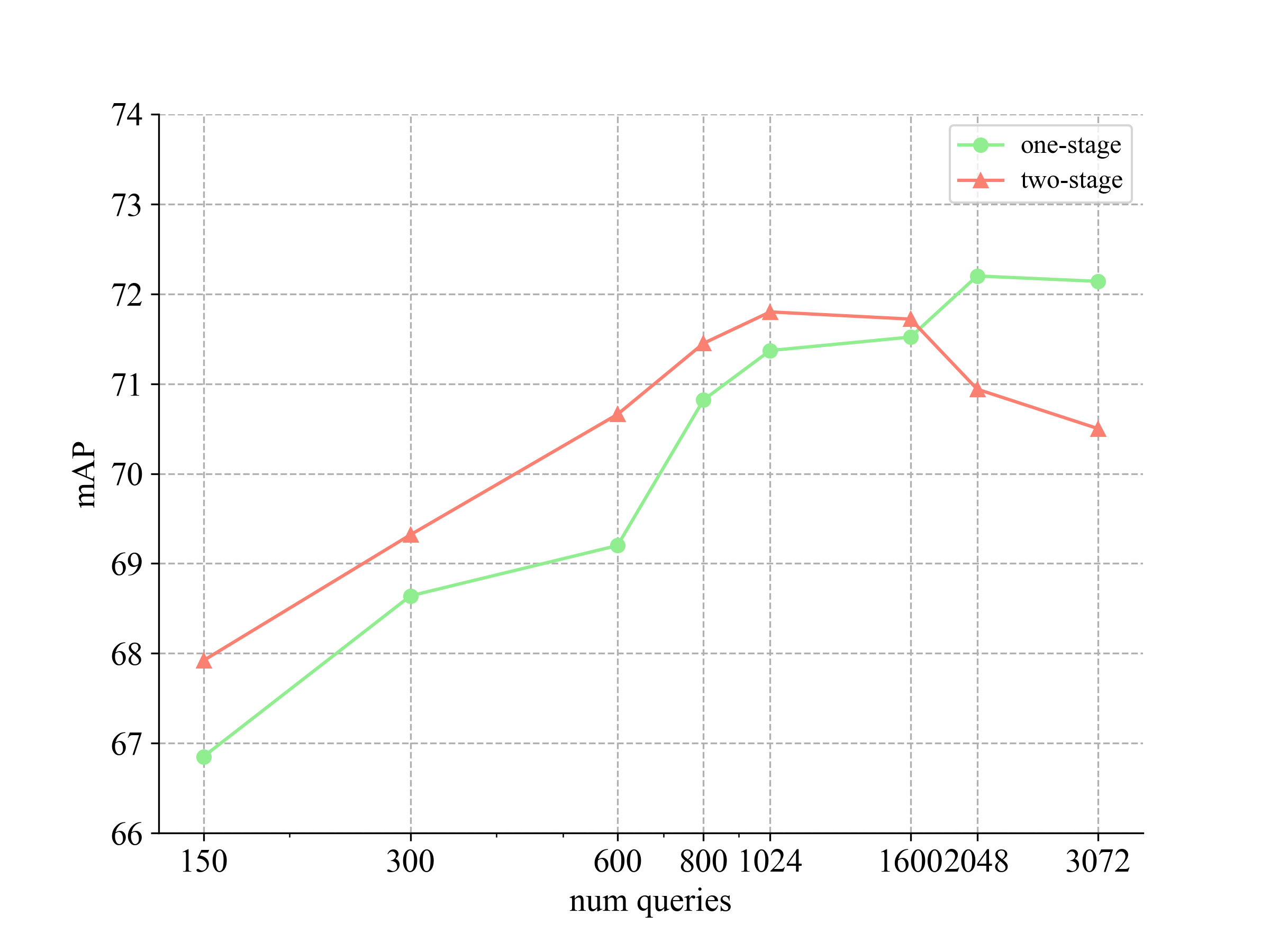}}
\captionsetup{font={small}}
\caption{(a) Effect of the offset range restriction on the deformable sampling offsets. (b) Influence of the number of queries.}
\label{fig:offset_nq}
\vspace{-16pt}
\end{figure}

\textbf{Effect of Range Restriction}.
RotaTR restricts the range of the sampling locations. By default, the sampling points are restricted to be inside the reference anchor. While considering that the reference anchor may not be accurate, we introduce the parameter $\alpha$ to control the sampling range. To verify the effect of the sampling range on the performance, we experiment on the offset range factor $\alpha$. The results are shown in Figure-\ref{fig:offset_nq}(a). 
It is seen that when no restriction, the performance only gets 69.2\% mAP. 
The experiment results generally follow the pattern that the larger range, the lower performance. The effect of the range restriction can also be validated through the quality plot in Figure-\ref{fig:pts_plot}, in which the sampled points of the deformable attention are plot. It is seen that when no range restriction is applied, the sampled points are not strictly aligned with the target instance. 
% figure, no restrict vs restrict
% \begin{figure}
%     \centering
%     \includegraphics[width=0.8\linewidth]{pics/offset.png}
%     \caption{Effect of the offset range restriction on the deformable sampling offsets.}
%     \label{fig:offset_bar}
% \end{figure}

% \textbf{Influence of the number of queries.} 
% \begin{figure}
%     \centering
%     \includegraphics[width=0.8\linewidth]{pics/numquery.png}
%     \caption{Influence of the number of queries.}
%     \label{fig:nq}
% \end{figure}

% \begin{figure}[H]
% \centering 
% \begin{minipage}[b]{0.225\textwidth}
% \centering 
% \includegraphics[width=1.0\textwidth]{pics/offset.png} 
% \subcaption{Effect of the offset range restriction on the deformable sampling offsets.}
% \label{fig:offset_bar}
% \end{minipage}
% \begin{minipage}[b]{0.225\textwidth} 
% \centering 
% \includegraphics[width=1.0\textwidth]{pics/numquery.png}
% \subcaption{Influence of the number of queries.}
% \label{fig:nq}
% \end{minipage}
% \caption{main}
% \end{figure}

\textbf{Influence of Number of Queries. }
% In this experiment, we compare the influence of changing the number of queries to the model performance in the DOTA-v1.0 validation set. 
The experiment for query number ablation is conducted for both one-stage and two-stage forms of RotaTR. The results are shown in Figure-\ref{fig:offset_nq}(b). Considering that the most number of instances per cropped patch is about 1,000, we think that setting the number of queries above 1,000 is appropriate. For the one-stage form, the increase of queries consistently leads to better performance. While for the two-stage form, the excess of queries will result in a performance drop. We argue that the essential influence is brought by the number of anchors. For the one-stage form, the number of anchors is simply defined by the randomly initialized queries. While the two-stage form makes use of every pixel of the feature maps as anchors and the number of queries only influences the top-K candidates' selection of them. So, too many candidates may result in a large number of redundant proposals input to the decoder, which leads to the performance drop.

\textbf{Effect of Label Assignment.}
In this experiment, we explore the effect of the label assignment of the alignment module. 
% The experiment is conducted for both one-stage and two-stage forms of RotaTR. 
The label assignment method for the decoder part is kept unchanged. We only vary the label assignment for the first stage (feature alignment module and the encoder output) and the label assignment methods are chosen from the original one-to-one matching (O2O) scheme and the ATSS (O2M) scheme. The results are shown in Table-\ref{tab:label_assign}. It is seen that the one-stage and two-stage form favors different assignment mechanisms. For the one-stage form, the O2M mechanism for the encoder output will result in better performance. For the two-stage form, only the O2O mechanism should be selected as the O2M will result in a huge performance decline (-7.61\%AP). We argue that the two-stage form highly depends on good proposal generation, while the O2M label assignment will bring a large number of redundant candidates, thus deteriorating the decoding process.

\begin{table}[]
\begin{tabular}{c|c|ccc}
\hline
Method                     & Lable Assign & AP50  & AP75  & AP50:95 \\ \hline
\multirow{2}{*}{One-Stage} & O2O          & 70.67 & 47.48 & 44.47   \\
                           & O2M          & \textbf{72.01} & \textbf{49.02} & \textbf{45.11}  \\ \hline
\multirow{2}{*}{Two-Stage} & O2O          & \textbf{71.92} & \textbf{48.01} & \textbf{44.37}   \\
                           & O2M          & 64.31 & 43.25 & 38.02   \\ \hline
\end{tabular}
\captionsetup{font={small}}
\caption{Effect of label assignment. The repsults are reported on DOTA-v1.0 validation set. O2O is the short for the one to one matching and O2M is the short for one to many matching.}
\label{tab:label_assign}
\vspace{-12pt}
\end{table}

\section{Conclusion}
This paper proposes RotaTR as an extension from the DETR to the oriented detection. We conduct experiments on several challenging oriented object detection benchmarks including the aerial object detection, ship detection and scene text detection. Experimental results show that our method has competitive accuracy to the state of the art. 

%%%%%%%%% REFERENCES
{\small
\bibliographystyle{ieee_fullname}
\bibliography{egbib}
}

\end{document}